\documentclass{article}

\usepackage{fullpage, epsf, setspace, latexsym, cite}
\usepackage{xspace, amsmath, amssymb, subfigure, hyperref}
\usepackage{bm}
\usepackage{graphicx}
\usepackage{color}
\usepackage{url}

\def\QED{\mbox{\rule[0pt]{1.5ex}{1.5ex}}}
\def\endproof{\hspace*{\fill}~\QED\par\endtrivlist\unskip}

\def\bf0{{{\bm 0}}}
\def\bf1{{{\bm 1}}}

\def\bfu{{{\mathbf u}}}
\def\bfU{{{\mathbf U}}}

\def\bfw{{{\mathbf w}}}
\def\bfW{{{\mathbf W}}}
\def\bfx{{{\mathbf x}}}
\def\bfX{{{\mathbf X}}}
\def\bfy{{{\mathbf y}}}
\def\bfY{{{\mathbf Y}}}

\def\E{\ensuremath{{\mathbb E}}\xspace}

\def\R{\ensuremath{{\mathbb R}}\xspace}

\def\argmax{\ensuremath{\mbox{argmax}}}
\def\argmin{\ensuremath{\mbox{argmin}}}

\title{An Optimal Control View of Adversarial Machine Learning}
\author{Xiaojin Zhu\\Department of Computer Sciences, University of Wisconsin-Madison}
\date{}
\begin{document}
\maketitle

\begin{abstract}
I describe an optimal control view of adversarial machine learning, where the dynamical system is the machine learner, the input are adversarial actions, and the control costs are defined by the adversary's goals to do harm and be hard to detect.
This view encompasses many types of adversarial machine learning, including test-item attacks, training-data poisoning, and adversarial reward shaping.
The view encourages adversarial machine learning researcher to utilize advances in control theory and reinforcement learning.
\end{abstract}

\section{Adversarial Machine Learning is not Machine Learning}
Machine learning has its mathematical foundation in concentration inequalities.
This is a consequence of the independent and identically-distributed (i.i.d.) data assumption.
In contrast, I suggest that adversarial machine learning may adopt optimal control as its mathematical foundation~\cite{Bertsekas2017Dynamic,todorov2006optimal}.
There are telltale signs: adversarial attacks tend to be subtle and have peculiar non-i.i.d. structures -- as control input might be.

\section{Optimal Control}
I will focus on deterministic discrete-time optimal control because it matches many existing adversarial attacks. 
Extensions to stochastic and continuous control are relevant to adversarial machine learning, too.
The system to be controlled is called the plant, which is defined by the system dynamics:
\begin{equation}
\bfx_{t+1} = f(\bfx_t, \bfu_t)
\label{eq:dynamics}
\end{equation}
where $\bfx_t \in \bfX_t$ is the state of the system,
$\bfu_t \in \bfU_t$ is the control input, and $\bfU_t$ is the control constraint set.
The function $f$ defines the evolution of state under external control. 
The time index $t$ ranges from 0 to $T-1$, and the time horizon $T$ can be finite or infinite.
The quality of control is specified by the running cost:
\begin{equation}
g_t(\bfx_t, \bfu_t)
\end{equation}
which defines the step-by-step control cost,
and the terminal cost for finite horizon:
\begin{equation}
g_T(\bfx_T)
\end{equation}
which defines the quality of the final state.
The optimal control problem is to find control inputs $\bfu_0 \ldots \bfu_{T-1}$ in order to minimize the objective:
\begin{eqnarray}
\min_{\bfu_0 \ldots \bfu_{T-1}} && g_T(\bfx_T) + \sum_{t=0}^{T-1} g_t(\bfx_t, \bfu_t) \label{eq:optimization} \\
\mbox{s.t.} &&  \bfx_{t+1} = f(\bfx_t, \bfu_t), \; \bfu_t \in \bfU_t, \; \forall t  \nonumber\\
	&& \bfx_0 \mbox{ given} \nonumber
\end{eqnarray}
More generally, the controller aims to find control policies $\phi_t(\bfx_t) = \bfu_t$, namely functions that map observed states to inputs.
In optimal control the dynamics $f$ is known to the controller.
There are two styles of solutions: dynamic programming and Pontryagin minimum principle~\cite{liberzon,athansfalb,friesz2010dynamic}.
When $f$ is not fully known, the problem becomes either robust control where control is carried out in a minimax fashion to accommodate the worst case dynamics~\cite{Zhu2017NoLearner}, or reinforcement learning where the controller probes the dynamics~\cite{2018arXiv180609460R}.

\section{Adversarial Machine Learning as Control}
Now let us translate adversarial machine learning into a control formulation.
Adversarial machine learning studies vulnerability throughout the learning pipeline~\cite{vorobeychik2018adversarial,joseph2018adversarial,DBLP:journals/corr/abs-1712-03141,lowd2005adversarial}.
As examples, I present 
training-data poisoning, 
test-time attacks, 
and adversarial reward shaping below.
In all cases, the adversary attempts to control the machine learning system, and the control costs reflect the adversary's desire to do harm and be hard to detect.

Unfortunately, the notations from the control community and the machine learning community clash.
For example, $\bfx$ denotes the state in control but the feature vector in machine learning.
I will use the machine learning convention below.

\subsection{Training-Data Poisoning}
In training-data poisoning the adversary can modify the training data.  The machine learner then trains a ``wrong'' model from the poisoned data.  The adversary's goal is for the ``wrong'' model to be useful for some nefarious purpose. 
I use supervised learning for illustration.

\subsubsection{Batch Learner}
At this point, it becomes useful to distinguish batch learning and sequential (online) learning.
If the machine learner performs batch learning, then the adversary has a degenerate \emph{one-step} control problem.
One-step control has not been the focus of the control community and there may not be ample algorithmic solutions to borrow from.
Still, it is illustrative to pose batch training set poisoning as a control problem. 
I use Support Vector Machine (SVM) with a batch training set as an example below:
\begin{itemize}
\item
The state is the learner's model $h: \bfX \mapsto \bfY$.  For instance, for SVM $h$ is the classifier parametrized by a weight vector $\bfw$.
I will use $h$ and $\bfw$ interchangeably.
\item
The control $\bfu_0$ is a whole training set, for instance $\bfu_0 = \{(\bfx_i, y_i)\}_{1:n}$.
\item
The control constraint set $\bfU_0$ consists of training sets available to the adversary; if the adversary can arbitrary modify a training set for supervised learning (including changing features and labels, inserting and deleting items), this could be $\bfU_0 = \cup_{n=0}^\infty (\bfX \times \bfY)^n$, namely all training sets of all sizes.
This is a large control space.
\item
The system dynamics~\eqref{eq:dynamics} is defined by the learner's learning algorithm.
For the SVM learner, this would be empirical risk minimization with hinge loss $\ell()$ and a regularizer:
\begin{equation} 
\bfw_1 = f(\bfu_0) \in \argmin_\bfw \sum_{i=1}^n \ell(\bfw, \bfx_i, y_i) + \lambda \|\bfw\|^2.
\label{eq:hinge}
\end{equation} 
The batch SVM does not need an initial weight $\bfw_0$.
The adversary has full knowledge of the dynamics $f()$ if it knows the form~\eqref{eq:hinge}, $\ell()$, and the value of $\lambda$.
\item The time horizon $T=1$.
\item The adversary's running cost $g_0(\bfu_0)$ measures the poisoning effort in preparing the training set $\bfu_0$.
This is typically defined with respect to a given ``clean'' data set $\tilde \bfu$ before poisoning in the form of
\begin{equation} 
g_0(\bfu_0) = \mathrm{distance}(\bfu_0, \tilde \bfu).
\end{equation} 
The running cost is domain dependent.
For example, the distance function may count the number of modified training items; or sum up the Euclidean distance of changes in feature vectors.
\item The adversary's terminal cost $g_1(\bfw_1)$ measures the lack of intended harm.
The terminal cost is also domain dependent.
For example:
  \begin{itemize}
  \item If the adversary must force the learner into exactly arriving at some target model $\bfw^*$, then $g_1(\bfw_1) = \mathbb I_\infty[\bfw_1 \neq \bfw^*]$.
Here $\mathbb I_y[z]=y$ if $z$ is true and 0 otherwise, which acts as a hard constraint.
  \item If the adversary only needs the learner to get near $\bfw^*$ then $g_1(\bfw_1) = \|\bfw_1 - \bfw^*\|$ for some norm.
  \item If the adversary wants to ensure that a specific future item $\bfx^*$ is classified $\epsilon$-confidently as positive, it can use 
$g_1(\bfw_1) = \mathbb I_\infty[\bfw_1 \notin \bfW^*]$ with the target set $\bfW^*=\{\bfw: \bfw^\top \bfx^* \ge \epsilon\}$.
More generally, $\bfW^*$ can be a polytope defined by multiple future classification constraints.
  \end{itemize}
With these definitions, the adversary's one-step control problem~\eqref{eq:optimization} specializes to
\begin{eqnarray}
\min_{\bfu_0} && g_1(\bfw_1) + g_0(\bfw_0, \bfu_0) \\
\mbox{s.t.} &&  \bfw_{1} = f(\bfw_0, \bfu_0)  \nonumber
\end{eqnarray}
\end{itemize}
Unsurprisingly, the adversary's one-step control problem is equivalent to a Stackelberg game and bi-level optimization (the lower level optimization is hidden in $f$), a well-known formulation for training-data poisoning~\cite{Mei2015Machine,jagielski2018manipulating}.

\subsubsection{Sequential Learner}
The adversary performs classic discrete-time control if the learner is sequential:
\begin{itemize}
\item The learner starts from an initial model $\bfw_0$, which is the initial state.
\item The control input at time $t$ is $\bfu_t=(\bfx_t, y_t)$, namely the $t^{th}$ training item for $t=0,1,\ldots$ 
\item The dynamics is the sequential update algorithm of the learner.  For example, the learner may perform one step of gradient descent:
\begin{equation}
\bfw_{t+1} = f(\bfw_t, \bfu_t) = \bfw_t - \eta_t \nabla \ell(\bfw_t, \bfx_t, y_t).
\end{equation}
\item The adversary's running cost $g_t(\bfw_t, \bfu_t)$ typically measures the effort of preparing $\bfu_t$.  For example, 
it could measure the magnitude of change $\|\bfu_t - \tilde \bfu_t\|$ with respect to a ``clean'' reference training sequence $\tilde \bfu$.
Or it could be the constant 1 which reflects the desire to have a short control sequence. 
\item The adversary's terminal cost $g_T(\bfw_T)$ is the same as in the batch case.
\end{itemize}
The problem~\eqref{eq:optimization} then produces the optimal training sequence poisoning.
Earlier attempts on sequential teaching can be found in~\cite{liu2017iterative,DBLP:conf/icml/LiuDLLRS18,Alfeld2016Data}.

\subsection{Test-Time Attack}
Test-time attack differs from training-data poisoning in that a machine learning model $h: \bfX \mapsto \bfY$ is already-trained and given.
Also given is a ``test item'' $\bfx$.
There are several variants of test-time attacks, I use the following one for illustration:
The adversary seeks to minimally perturb $\bfx$ into $\bfx'$ such that the machine learning model classifies $\bfx$ and $\bfx'$ differently. 
That is,
\begin{eqnarray}
\label{eq:testtimeattack}
\min_{\bfx'} && \mathrm{distance}(\bfx, \bfx') \\
\mbox{s.t.} && h(\bfx) \neq h(\bfy). \nonumber
\end{eqnarray}
The distance function is domain-dependent, though in practice the adversary often uses a mathematically convenient surrogate such as some $p$-norm $\|\bfx-\bfx'\|_p$. 

One way to formulate test-time attack as optimal control is to treat the test-item itself as the state, and the adversarial actions as control input.
Let us first look at the popular example of test-time attack against image classification:
\begin{itemize}
\item Let the initial state $\bfx_0 = \bfx$ be the clean image.
\item The adversary's control input $\bfu_0$ is the vector of pixel value changes.
\item The control constraint set is $\bfU_0=\{\bfu: \bfx_0 + \bfu \in [0,1]^d\}$ to ensure that the modified image has valid pixel values (assumed to be normalized in $[0,1]$).
\item The dynamical system is trivially vector addition: $\bfx_1 = f(\bfx_0, \bfu_0) = \bfx_0+\bfu_0$.
\item The adversary's running cost is $g_0(\bfx_0, \bfu_0) = \mathrm{distance}(\bfx_0, \bfx_1)$.
\item The adversary's terminal cost is $g_1(\bfx_1) = \mathbb I_\infty[h(\bfx_1)=h(\bfx_0)]$.
Note the machine learning model $h$ is only used to define the hard constraint terminal cost; $h$ itself is not modified.
\end{itemize}
With these definitions this is a one-step control problem~\eqref{eq:optimization} that is equivalent to the test-time attack problem~\eqref{eq:testtimeattack}. 

This control view on test-time attack is more interesting when the adversary's actions are sequential $\bfU_0, \bfU_1, \ldots$, and the system dynamics render the action sequence non-commutative.
The adversary's running cost $g_t$ then measures the effort in performing the action at step $t$.
One limitation of the optimal control view is that the action cost is assumed to be additive over the steps.

\subsection{Defense Against Test-Time Attack by Adversarial Training}
Some defense strategies can be viewed as optimal control, too.
One defense against test-time attack is to require the learned model $h$ to have the large-margin property with respect to a training set.
Let $(\bfx, y)$ be any training item, and $\epsilon$ a margin parameter.  Then the large-margin property states that the decision boundary induced by $h$ should not pass $\epsilon$-close to $(\bfx,y)$:
\begin{equation}
\forall \bfx': \left(\|\bfx'-\bfx\|_p \le \epsilon \right) \Rightarrow h(\bfx')=y.
\label{eq:margin}
\end{equation}
This is an uncountable number of constraints.  It is relatively easy to enforce for linear learners such as SVMs, but impractical otherwise.

Adversarial training can be viewed as a heuristic to approximate the uncountable constraint~\eqref{eq:margin} with a finite number of active constraints:
one performs test-time attack against the current $h$ from $\bfx$ to find an adversarial item $\bfx^{(1)}$, such that $\|\bfx^{(1)}-\bfx\|_p \le \epsilon$ but $h(\bfx^{(1)})\neq y$.
Instead of adding a single constraint $h(\bfx^{(1)})= y$, an additional training item $(\bfx^{(1)}, y)$ is then added to the training set.
The machine learning algorithm learns a different $h$, with the hope (but not constraining) that $h(\bfx^{(1)})= y$.
This process repeats for $k$ iteration, resulting in $k$ additional training items $(\bfx^{(i)}, y)$ for $i=1\ldots k$.

It should be clear that such defense is similar to training-data poisoning, in that the defender uses data to modify the learned model.
This is especially interesting when the learner performs sequential updates.
One way to formulate adversarial training defense as control is the following:
\begin{itemize}
\item The state is the model $h_t$.  Initially $h_0$ can be the model trained on the original training data.
\item The control input $\bfu_t=(\bfx_t, y_t)$ is an additional training item with the trivial constraint set $\bfU_t=\bfX \times \bfy$.
\item The dynamics $h_{t+1} = f(h_t, \bfu_t)$ is one-step update of the model, e.g. by back-propagation.
\item The defender's running cost $g_t(h_t, \bfu_t)$ can simply be 1 to reflect the desire for less effort (the running cost sums up to $k$). 
\item The defender's terminal cost $g_T(h_T)$ penalizes small margin of the final model $h_T$ with respect to the original training data.
\end{itemize}
Of course, the resulting control problem~\eqref{eq:optimization} does not directly utilize adversarial examples.
One way to incorporate them is to restrict $\bfU_t$ to a set of adversarial examples found by invoking test-time attackers on $h_t$, similar to the heuristic in~\cite{cai2018curriculum}.
These adversarial examples do not even need to be successful attacks.

\subsection{Adversarial Reward Shaping}
When adversarial attacks are applied to sequential decision makers such as multi-armed bandits or reinforcement learning agents, a typical attack goal is to force the latter to learn a wrong policy useful to the adversary.
The adversary may do so by manipulating the rewards and the states experienced by the learner~\cite{46154,Jun2018Adversarial}.


To simplify the exposition, I focus on adversarial reward shaping against stochastic multi-armed bandit, because this does not involve deception through perceived states.
To review, in stochastic multi-armed bandit the learner at iteration $t$ chooses one of $k$ arms, denoted by $I_t \in [k]$, to pull according to some strategy~\cite{bubeck2012regret}.
For example, the $(\alpha, \psi)$-Upper Confidence Bound (UCB) strategy chooses the arm
\begin{equation}
I_t \in \argmax_{i\in[k]} \hat\mu_{i, T_i(t-1)} + {\psi^*}^{-1}\left({ \alpha \log t \over T_i(t-1) } \right)
\label{eq:banditarm}
\end{equation}
where $T_i(t-1)$ is the number of times arm $i$ has been pulled up to time $t-1$, $\hat\mu_{i, T_i(t-1)}$ is the empirical mean of arm $i$ so far, and $\psi^*$ is the dual of a convex function $\psi$.
The environment generates a stochastic reward $r_{I_t} \sim \nu_{I_t}$.
The learner updates its estimate of the pulled arm: 
\begin{equation}
\hat\mu_{I_t, T_{I_t}(t)}
= {
\hat\mu_{I_t, T_{I_t}(t-1)} T_{I_t}(t-1) + r_{I_t}
\over
T_{I_t}(t-1) + 1
}
\label{eq:banditupdate}
\end{equation}
which in turn affects which arm it will pull in the next iteration.
The learner's goal is to minimize the pseudo-regret $T \mu^{\max} - \E \sum_{t=1}^T \mu_{I_t}$ where $\mu_i = \E \nu_i$ and $\mu^{\max} = \max_{i\in[k]} \mu_i$.
Stochastic multi-armed bandit strategies offer upper bounds on the pseudo-regret.

With adversarial reward shaping, an adversary fully observes the bandit.
The adversary intercepts the environmental reward $r_{I_t}$ in each iteration, and may choose to modify (``shape'') the reward into
$$r_{I_t} + u_t$$
with some $u_t \in \R$ before sending the modified reward to the learner.
The adversary's goal is to use minimal reward shaping to force the learner into performing specific wrong actions.
For example, the adversary may want the learner to frequently pull a particular target arm $i^* \in [k]$.
It should be noted that the adversary's goal may not be the exact opposite of the learner's goal: the target arm $i^*$ is not necessarily the one with the worst mean reward, and the adversary may not seek pseudo-regret maximization.

Adversarial reward shaping can be formulated as stochastic optimal control:
\begin{itemize}
\item The state $s_t$, now called control state to avoid confusion with the Markov Decision Process states experienced by an reinforcement learning agent, consists of the sufficient statistic tuple at time $t$:
$$s_t = (T_1(t-1), \hat\mu_{1, T_1(t-1)}, \ldots, T_k(t-1), \hat\mu_{k, T_k(t-1)}, I_t).$$ 
\item The control input is $u_t \in \bfU_t$ with $\bfU_t=\R$ in the unconstrained shaping case, or the appropriate $\bfU_t$ if the rewards must be binary, for example.
\item The dynamics $s_{t+1}=f(s_t, u_t)$ is straightforward via  empirical mean update~\eqref{eq:banditupdate}, $T_{I_t}$ increment,  and new arm choice~\eqref{eq:banditarm}.
\item The adversary's running cost $g_t(s_t, u_t)$ reflects shaping effort and target arm achievement in iteration $t$.
For instance, 
\begin{equation}
g_t(s_t, u_t) = u_t^2 + \mathbb I_\lambda[I_t \neq i^*].
\end{equation}
where $\lambda > 0$ is a trade off parameter.
\item There is not necessarily a time horizon $T$ or a terminal cost $g_T(s_T)$.
\end{itemize}
The control state is stochastic due to the stochastic reward $r_{I_t}$ entering through~\eqref{eq:banditupdate}.

\section{Advantages of the Optimal Control View}

There are a number of potential benefits in taking the optimal control view:
\begin{itemize}
\item
It offers a unified conceptual framework for adversarial machine learning;
\item
The optimal control literature provides efficient solutions when the dynamics $f$ is known and one can take the continuous limit to solve the differential equations \cite{2018arXiv181006175L};
\item
Reinforcement learning, either model-based with coarse system identification or model-free policy iteration, allows approximate optimal control when $f$ is unknown, as long as the adversary can probe the dynamics~\cite{Fan2018Learning,pmlr-v80-dai18b};
\item
A generic defense strategy may be to limit the controllability the adversary has over the learner.
\item
I mention in passing that the optimal control view applies equally to machine teaching~\cite{Zhu2018Overview,Zhu2015Machine}, and thus extends to the application of personalized education~\cite{Sen2018Machine,Patil2014Optimal}.
\end{itemize}

I need to point out some limitations:
\begin{itemize}
\item
Having a unified optimal control view does not automatically produce efficient solutions to the control problem~\eqref{eq:optimization}.
For adversarial machine learning applications the dynamics $f$ is usually highly nonlinear and complex.
Furthermore, in graybox and blackbox attack settings $f$ is not fully known to the attacker.
They affect the complexity in finding an optimal control.
\item 
The adversarial learning setting is largely non-game theoretic, though there are exceptions~\cite{bruckner2011stackelberg,pmlr-v38-li15a}.
\end{itemize}
These problems call for future research from both machine learning and control communities.

\textbf{Acknowledgments.} I acknowledge funding NSF 1837132, 1545481, 1704117, 1623605, 1561512, and the MADLab AF Center of Excellence FA9550-18-1-0166.

\bibliographystyle{plain}
\bibliography{amloc}
\end{document}